\documentclass[lettersize,journal]{IEEEtran}
\usepackage{amsmath,amsfonts}
\usepackage{array}
\usepackage[caption=false,font=normalsize,labelfont=sf,textfont=sf]{subfig}
\usepackage{textcomp}
\usepackage{stfloats}
\usepackage{url}
\usepackage{verbatim}
\usepackage{graphicx}
\usepackage{cite}
\usepackage[whole]{bxcjkjatype}
\usepackage{xcolor}
\usepackage{amssymb,amsbsy}
\usepackage{graphicx}
\usepackage{mathtools}
\usepackage{algorithm}
\usepackage{algpseudocode}
\usepackage{xspace}

\usepackage{booktabs,multirow,siunitx,makecell}
\usepackage[table]{xcolor}
\newcommand{\bfhl}[1]{\cellcolor{red!12}\textbf{#1}}
\newcommand{\hl}[1]{\cellcolor{red!12}#1}

\newcommand{\etal}{\emph{et~al.}\xspace}
\usepackage[dvipsnames]{xcolor}

\renewcommand{\vec}[1]{\boldsymbol{#1}}

\begin{document}

\title{Single-Image Depth from Defocus with Coded Aperture and Diffusion Posterior Sampling}

\author{Hodaka Kawachi, Jos\'{e} Reinaldo Cunha Santos A. V. Silva Neto, Yasushi Yagi,~\IEEEmembership{Senior Member, IEEE}, Hajime Nagahara and Tomoya Nakamura
\thanks{
This work was supported by the JST FOREST Program (Grant Number JPMJFR206K); JSPS KAKENHI (Grant Numbers JP23H05490, JP24KJ1660); and the Konica Minolta Hikari-Mirai Encouragement Fund administered by The Optical Society of Japan (funded by the Konica Minolta Science and Technology Foundation).

Hodaka Kawachi and Jos\'{e} Reinaldo Cunha Santos A. V. Silva Neto  
are with the SANKEN, The University of Osaka, 5-1 Mihogaoka, Ibaraki, Osaka, 567-0047, Japan (e-mail: kawachi@im.sanken.osaka-u.ac.jp; vieira@im.sanken.osaka-u.ac.jp).
Yasushi Yagi and Hajime Nagahara are with the D3 center, The University of Osaka, 5-1 Mihogaoka, Ibaraki, Osaka 567-0047, Japan (email: yagi@am.sanken.osaka-u.ac.jp; nagahara@ids.osaka-u.ac.jp).
Tomoya Nakamura is with the Graduate School of Engineering Science, The University of Osaka, 
1-3 Machikaneyama, Toyonaka, Osaka 560-8531, Japan (e-mail: t.nakamura.opt@osaka-u.ac.jp).
}}

\markboth{Journal of \LaTeX\ Class Files,~Vol.~14, No.~8, August~2021}%
{Shell \MakeLowercase{\textit{et al.}}: A Sample Article Using IEEEtran.cls for IEEE Journals}


\maketitle

\begin{abstract}
We propose a single-snapshot depth-from-defocus (DFD) reconstruction method for coded-aperture imaging that replaces hand-crafted priors with a learned diffusion prior used purely as regularization. Our optimization framework enforces measurement consistency via a differentiable forward model while guiding solutions with the diffusion prior in the denoised image domain, yielding higher accuracy and stability than classical optimization. Unlike U-Net–style regressors, our approach requires no paired defocus–RGBD training data and does not tie training to a specific camera configuration. Experiments on comprehensive simulations and a prototype camera demonstrate consistently strong RGBD reconstructions across noise levels, outperforming both U-Net baselines and a classical coded-aperture DFD method.
\end{abstract}


\section{Introduction}
The acquisition of scene geometry in the form of RGBD data remains a fundamental problem in computer vision. Although high-quality depth sensors, such as LiDAR and indirect time-of-flight cameras, have become widely available, they are costly and impose constraints on placement and synchronization~\cite{Kawachi2023a,foix2011lock}. Passive stereo rigs are a mature alternative, but they require at least two synchronized cameras with a suitable baseline, careful calibration, and sufficient scene texture/lighting to avoid ambiguities~\cite{Scharstein2002,Szeliski2010}. Consequently, monocular depth estimation---with minimal hardware---continues to be an active field of research.

Recovering a 3D structure from a single RGB image is inherently ill-posed. Hence, methods that leverage auxiliary cues at the time of image capture have attracted considerable attention. For instance, the coded-aperture depth-from-defocus~(DFD) approach developed by Levin \textit{et~al.}~\cite{Levin2007} exploits the defocus blur recorded at imaging to improve the depth accuracy. This concept has inspired extensive follow-up work, including ours. Notably, it was the first to demonstrate depth reconstruction from a single snapshot---introducing a coded aperture helps disentangle scene texture from defocus components, thereby mitigating the ill-posedness.

\begin{figure}
    \centering
    \includegraphics[width=1.0\linewidth]{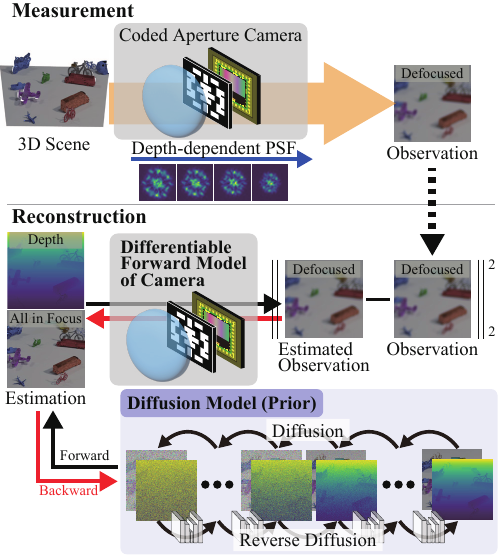}
    \caption{Concept of our single-image depth-from-defocus (DFD) with diffusion posterior sampling tailored for DFD (DFD-DPS).
    In the measurement process, a 3D scene is observed as a single 2D image affected by depth-dependent coded point spread functions (PSFs).
    In the reconstruction process, an all-in-focus image and depth map are reconstructed from a single observation image by minimizing the error between the physical and estimated observation obtained through the differential forward model.
    To solve this ill-posed problem, the diffusion model is integrated into the reconstruction pipeline as a strong prior of natural images.}
    \label{fig:abstract}
\end{figure}

The approach of Levin \textit{et~al.} adopts a classical optimization framework that combines a data-fidelity term with a prior (regularization) term to stabilize the solution. Such optimization-based methods avoid the need for large labeled datasets and work with calibrated cameras. However, they rely on hand-crafted priors, and thus have intrinsic accuracy limits, especially in low-texture regions or under strong blur.

With the advent of deep learning, regression-based DFD methods that learn a direct mapping from defocused images to depth---such as end-to-end U-Net---have surpassed the accuracy of classical optimization-based approaches. However, because regression-based DFD approximates the inverse mapping rather than solving it explicitly, these methods require camera parameters and optical conditions to remain consistent between training and inference. They also demand large paired datasets of defocused images aligned with RGB-D ground truth. In practice, such pairs are difficult to acquire and are often synthesized. The resulting sim-to-real gap---including point spread function (PSF) mismatch, noise characteristics, and differences in the camera's image signal processing pipeline---can markedly degrade performance on real image captures, limiting robustness and generality.

To improve depth accuracy in real settings, we return to joint optimization, but replace hand-crafted priors with a high-capacity diffusion prior. Specifically, we couple a differentiable coded-aperture forward model in the data-fidelity term with a diffusion model trained for RGBD generation, which is used solely as the regularizer. This design decouples camera calibration from network training and enables flexible single-image DFD without paired defocus--RGBD data, while maintaining strong measurement consistency during inference.

In summary, we introduce a computational-imaging reconstruction framework that combines a calibrated coded-aperture forward model with a high-capacity diffusion prior, replacing hand-crafted priors and bypassing end-to-end U-Net regression. 
This design yields a method that is more stable under train--test mismatches (e.g., noise and optics) and achieves higher reconstruction accuracy than regression-based baselines (e.g., U-Net), while not requiring paired defocus--RGBD data. 
We validate the approach through both large-scale simulations and real-prototype experiments. The results demonstrate that the proposed method achieves accurate single-image RGBD recovery from coded-defocus measurements.

\section{Related Work}

\subsection{Snapshot DFD}
Classical DFD recovers depth information by analyzing the defocus blur, and often employs coded apertures or phase masks to improve identifiability \cite{Levin2007,Zhou2010,Haim2018}.
These methods address the problem of ill-posedness by 
(i) discretizing the depth volume and 
(ii) imposing handcrafted regularizers. 
Recent learning-based variants adopt U-Net–style regressors \cite{Tang2017,Gur2019,Wu2019,Baek2022,Chang2019}, but their end-to-end training couples the network to a \textbf{fixed} optical configuration (mask/lens/F-number/focus/PSF, noise, and image signal processing). As a result, the camera parameters used at inference must match those seen in training. Typically, performance degrades sharply under changes in optics, focus setting, or mask. 
Although several studies have explored the construction of custom paired datasets of RGBD and defocused images under their own optics, these collections are typically small and far less diverse than large public RGBD datasets. This highlights the need for reconstruction methods that can directly leverage large, readily available RGBD datasets---without paired defocus captures---while maintaining accuracy on real scenes.
Another line of work leverages Deep Image Prior (DIP)~\cite{Ulyanov2020}, which has been widely used in computational imaging \cite{Monakhova2021, Neto_etal_RadialOpt_2023_IeeeTci, Kawachi2023a}. 
In DIP, the structure of an untrained convolutional network itself acts as a powerful regularizer. 
Such methods are attractive because they do not require training data. 
However, whether DIP can be effectively applied to DFD remains an open question.
In the proposed method, we return to the optimization view of DFD and replace handcrafted priors with a high-capacity diffusion prior used purely for regularization, thereby avoiding the need for paired defocus--RGBD data and decoupling the learning process from camera calibration.

\subsection{Diffusion Probabilistic Models}
Diffusion probabilistic models (DPMs) generate images by reversing a Markovian noise process \cite{ho2020denoising,sohl2015deep}.
A neural network estimates the score
$s_\theta(\vec{x}_t,t)\!\approx\!\nabla_{\vec{x}_t}\log p_t(\vec{x}_t)$,
which guides samples toward high-density regions (see Sec.~\ref{sec:prelim:dpm}).
DDPM~\cite{ho2020denoising} uses stochastic reverse steps, whereas DDIM~\cite{song2020denoising} provides a deterministic variant.

Since 2020, improved training/sampling objectives and guidance have led to DPMs becoming the state-of-the-art approach in image synthesis and conditional generation
\cite{dhariwal2021diffusion,song2021scorebased,ho2022classifierfree}.
Efficiency advances, such as latent diffusion, have further enabled high-resolution deployment
\cite{rombach2022ldm}.
As a result, diffusion models now underpin applications from text-to-image generation and editing
\cite{nichol2022glide,saharia2022imagen,rombach2022ldm}
to image restoration and image-to-image translation (e.g., SR3 super-resolution, Palette)
\cite{saharia2021sr3,saharia2022palette}.
For inverse problems, diffusion priors provide powerful regularizers or sampling targets
\cite{song2022medical_inverse,chung2023diffusion},
which motivates our use of diffusion as a learned prior for DFD.

\subsection{Diffusion Posterior Sampling }
Diffusion posterior sampling (DPS)~\cite{chung2023diffusion,chung2023parallel} incorporates observation information $\vec{y}$ into the diffusion reverse process by steering the trajectory toward the posterior over noisy states. 
Specifically, it augments the standard diffusion drift (the score prior) with an observation-consistency correction derived from the measurement model. Because the likelihood in the noisy space is intractable, DPS estimates a clean image $\hat{\vec{x}}_0$ from $\vec{x}_t$ via a Tweedie-based approximation, and then uses a reconstruction objective (e.g., consistency under the forward operator $A$) to guide the updates (see Sec.~\ref{sec:prelim:dps} for details).

DPS has achieved excellent results on restoration tasks with strong pixel-wise correspondence between $\vec{y}$ and $\vec{x}$ (e.g., inpainting and super-resolution). In DFD, however, the forward operator discards substantial information and the blur-–depth relationship is only indirectly reflected in the image domain. As a result, the observation-driven correction applied in the noisy space can be too weak to enforce the measurement constraint, a limitation highlighted in our experiments and addressed with our $\vec{x}_0$-guided update.

\section{Preliminaries}\label{sec:prelim}
In this section, we summarize the necessary preliminaries and background concepts that will be used in describing the proposed method.

\subsection{Notation}
We denote clean images by bold variables $\vec{x}_0\!\in\!\mathbb{R}^{H\times W\times3}$
and their noisy counterparts at diffusion step $t$ by
$\vec{x}_t$ ($t=1,\dots,T$).
A bar over~$\alpha$ indicates the cumulative product,
i.e., $\bar\alpha_t=\prod_{k=1}^{t}\alpha_k$.
The measurement operator for depth‐dependent blur is $A$, and the observed image is $\vec{y}$. 
Here we write it as a matrix multiplication for notational simplicity, even though the operation cannot be exactly expressed in this form.
$\boldsymbol{\epsilon}$ denotes independent and identically distributed (i.i.d.) standard Gaussian noise.

\subsection{Diffusion Probabilistic Models}\label{sec:prelim:dpm}
\paragraph{Forward process.}
Given $\vec{x}_0\sim p_{\text{data}}$, the DDPM/VP--SDE forward (noise) process
admits the following closed-form marginal \cite{ho2020denoising}:
\begin{equation}\label{eq:q}
  q(\vec{x}_t \mid \vec{x}_0)
  \;=\;
  \mathcal{N}\!\bigl(
    \vec{x}_t\,;\,
    \sqrt{\bar\alpha_t}\,\vec{x}_0,\;
    (1-\bar\alpha_t)\,I
  \bigr).
\end{equation}

\paragraph{Reverse process.}
We parameterize the reverse dynamics with a \emph{noise-prediction network}
$\epsilon_\theta:(\vec{x}_t,t)\mapsto\hat{\boldsymbol{\epsilon}}$ that predicts the additive Gaussian noise in \eqref{eq:q}.
The network is trained by minimizing
\begin{align}
\mathcal{L}_\epsilon
&= \mathbb{E}_{t,\vec{x}_0,\boldsymbol{\epsilon}}
   \Bigl[ \left\lVert \epsilon_\theta(\vec{x}_t,t)-\boldsymbol{\epsilon}\right\rVert_2^2 \Bigr], \\
&\text{where }\boldsymbol{\epsilon}\sim\mathcal{N}(\mathbf{0},\mathbf{I}),\quad
  \vec{x}_t \text{ is given by \eqref{eq:q}}. \nonumber
\end{align}
During sampling (i.e., in the reverse process), the predicted noise yields an estimate of the \emph{score}
(i.e., the gradient of the log-density),
\begin{equation}
  s_\theta(\vec{x}_t,t)\;\approx\;\nabla_{\vec{x}_t}\log p_t(\vec{x}_t)
  \;\approx\; -\frac{1}{\sqrt{1-\bar\alpha_t}}\;\epsilon_\theta(\vec{x}_t,t),
\end{equation}
which is used to perform stochastic (DDPM) or deterministic (DDIM~\cite{song2020denoising}) reverse updates.

\subsection{Posterior Mean via Tweedie's Formula}\label{sec:prelim:tweedie}
For the Gaussian forward process in \eqref{eq:q} (DDPM/VP--SDE),
Tweedie's formula~\cite{efron2011tweedie,kim2021noise} gives the posterior mean
of the clean sample:
\begin{equation}\label{eq:tweedie}
  \hat{\vec{x}}_0 := \mathbb{E}[\vec{x}_0 \mid \vec{x}_t]
  \;=\;
  \frac{1}{\sqrt{\bar\alpha_t}}
  \Bigl(\vec{x}_t + (1-\bar\alpha_t)\,\nabla_{\vec{x}_t}\log p_t(\vec{x}_t)\Bigr).
\end{equation}
In practice, the true score is replaced by its learned approximation
$s_\theta(\vec{x}_t,t)$, yielding
\begin{equation}
  \hat{\vec{x}}_0 \;\approx\;
  \frac{1}{\sqrt{\bar\alpha_t}}
  \Bigl(\vec{x}_t + (1-\bar\alpha_t)\,s_\theta(\vec{x}_t,t)\Bigr).
\end{equation}
Under the DDPM ``noise-prediction'' parameterization,
\begin{equation}
  s_\theta(\vec{x}_t,t) \;\approx\; -\,\frac{1}{\sqrt{1-\bar\alpha_t}}\,
  \epsilon_\theta(\vec{x}_t,t),
\end{equation}
which leads to the commonly used estimator
\begin{equation}
\label{eq:x_0_hat}
  \hat{\vec{x}}_0 \;\approx\;
  \frac{1}{\sqrt{\bar\alpha_t}}
  \Bigl(\vec{x}_t - \sqrt{1-\bar\alpha_t}\;\epsilon_\theta(\vec{x}_t,t)\Bigr).
\end{equation}
We will use $\hat{\vec{x}}_0=\hat{\vec{x}}_0(\vec{x}_t)$ as the denoised estimate in Sec.~\ref{sec:prelim:dps}.

\subsection{DPS}\label{sec:prelim:dps}
Let $p(\vec{y}\!\mid\!\vec{x}_0)$ be a differentiable measurement model (e.g., blur+noise).
DPS steers the trajectory toward the posterior $p(\vec{x}_t\!\mid\!\vec{y})$ by augmenting the base reverse update with an observation-consistency term:
\begin{equation}\label{eq:dps-grad}
  \vec{x}_{t-1} \;=\;
  \Phi_{\text{rev}}\!\bigl(\vec{x}_t;\,s_\theta\bigr)
  \;+\; \tau_t\,\nabla_{\vec{x}_t}\log p(\vec{y}\mid\vec{x}_t),
\end{equation}
where $\Phi_{\text{rev}}$ denotes the standard DDPM/DDIM reverse step and $\tau_t$ is a (possibly time-dependent) step size.
Because \(p(\vec{y}\mid\vec{x}_t)\) lacks a closed form,
\cite{chung2023diffusion} approximated it by marginalizing over
$\vec{x}_0$ and substituting the Tweedie posterior mean
$\hat{\vec{x}}_0$ from~\eqref{eq:x_0_hat}:
\begin{align}
  p(\vec{y}\mid\vec{x}_t)
    &\approx p(\vec{y}\mid\hat{\vec{x}}_0),\\
  \nabla_{\vec{x}_t}\log p(\vec{y}\mid\vec{x}_t)
    &\approx
      -\,\nabla_{\vec{x}_t}
      \bigl\lVert
          A\,\hat{\vec{x}}_0(\vec{x}_t)-\vec{y}
      \bigr\rVert^2,    \label{eq:dps-approx-grad}
\end{align}
providing the \emph{observation term} in~\eqref{eq:dps-grad}, while the score $s_\theta$ ensures that the trajectory remains on the data manifold.

\begin{figure}[t]
    \centering
    \includegraphics[width=\linewidth]{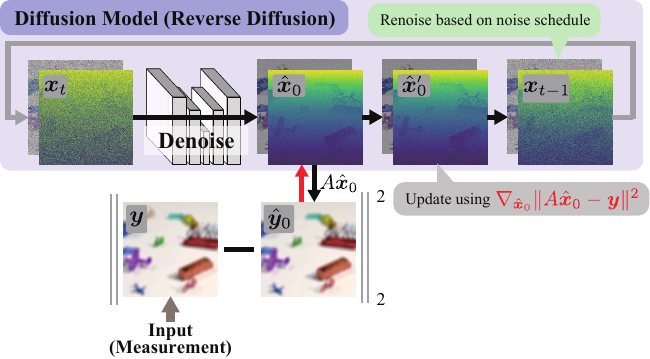}
    \caption{Pipeline of the proposed DFD\textendash DPS. At each step, we first denoise to obtain a provisional $\hat{\vec{x}}_0$. This is updated using the gradient of the observation error, and then re-noising is applied to advance the diffusion sampling process.}
    \label{fig:pipeline}
\end{figure}

\begin{figure}[t]
    \centering
    \includegraphics[width=\linewidth]{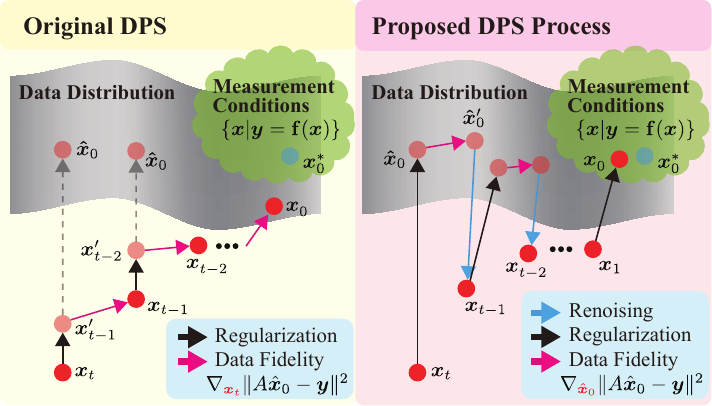}
    \caption{Generation process: original DPS vs.~our method. DPS enforces data fidelity by taking small-gradient steps in the noisy $\vec{x}_t$ space, where small moves translate ambiguously to changes in $\vec{x}_0$, especially under information-discarding operators (e.g., DFD). Hence, the constraint $A\vec{x}\approx\vec{y}$ is only weakly enforced. In contrast, we update the denoised estimate directly in $\vec{x}0$ using $\nabla{\vec{x}_0}\lVert A\vec{x}_0-\vec{y}\rVert^2$ before re-noising, which explicitly enforces the measurement and reduces hallucinations.}
    \label{fig:generation_process}
\end{figure}

\begin{algorithm}[t]
\caption{DFD--DPS}
\begin{algorithmic}[1]
\Require $N, \vec{y}, \{ \xi \}^{N}_{i=1}, \{ \tau \}^{N}_{i=1}, \{ \tilde{\sigma} \}^{N}_{i=1}$
\State $\vec{x}_{N-1} \sim \mathcal{N}(\vec{0}, \vec{I})$
\For{$i = N - 1$ \textbf{to} $0$}
    \State $\hat{\vec{s}} \leftarrow \vec{s}_\theta(\vec{x}_i, i)$
    \State $\hat{\vec{x}}_0 \leftarrow \frac{1}{\sqrt{\overline{\alpha}_i}} (\vec{x}_i + (1 - \overline{\alpha}_i) \hat{\vec{s}})$
    \State $\hat{\vec{x}}'_{0} \leftarrow \hat{\vec{x}}_{0} - \tau_i \nabla_{\hat{\vec{x}}_0} \Vert \vec{y} - f(\hat{\vec{x}}_0') \Vert_2^2$
    \State $\vec{z} \sim \mathcal{N}(\vec{0}, \vec{I})$
    \State $\vec{x}_{i-1} \leftarrow \frac{\sqrt{\overline{\alpha}_i}(1-\overline{\alpha}_{i-1})}{1-\overline{\alpha}_i} \vec{x}_i + \frac{\sqrt{\overline{\alpha}_{i-1}}\beta_i}{1-\overline{\alpha}_i} \hat{\vec{x}}'_0 + \tilde{\sigma}_i \vec{z}$
\EndFor
\State \Return $\vec{x}_0$
\end{algorithmic}
\label{algo:DPS}
\end{algorithm}

\section{Proposed Method}
This section details the proposed DFD\textendash DPS.

\subsection{DFD\textendash DPS}
\label{sec:DFD-DPS}
As reviewed in Sec.~\ref{sec:prelim:dps}, DPS~\cite{chung2023diffusion} uses a diffusion prior as a regularizer and augments the reverse update with an observation term, ensuring that the sample both follows the data distribution $p(\vec{x})$ and fits the measurement. Specifically, the observation term used in DPS is the gradient
\begin{equation}
    -\,\nabla_{\vec{x}_t}\,\bigl\|A\,\hat{\vec{x}}_0(\vec{x}_t)-\vec{y}\bigr\|^2,
\end{equation}
which indicates how to move $\vec{x}_t$ so that the denoised estimate $\hat{\vec{x}}_0$ better satisfies $A\hat{\vec{x}}_0 \approx \vec{y}$. 
By contrast, when $\vec{y}$ and $\vec{x}$ exhibit strong pixelwise correspondence (e.g., super‐resolution, inpainting, or Gaussian/motion deblurring), the gradient through $A$ is informative, and DPS reliably improves the measurement consistency (cf.\ \cite{chung2023diffusion}).
However, because $\vec{x}_t$ lives in a highly noisy space, small moves in $\vec{x}_t$ translate ambiguously to semantic changes in $\vec{x}_0$; this ambiguity becomes especially problematic when the forward operator $A$ discards substantial information, as in DFD (see Fig.~\ref{fig:generation_process}).

Motivated by DDIM~\cite{song2020denoising}, we incorporate the ``explicit $\vec{x}_0$ estimation + re‐noising'' structure into DPS~(Fig.~\ref{fig:pipeline}). At each step, we first obtain a provisional denoised image $\hat{\vec{x}}_0=\hat{\vec{x}}_0(\vec{x}_t)$ (e.g., via the Tweedie estimate in \eqref{eq:x_0_hat}), then update \emph{in the $\vec{x}_0$ space} to enforce the measurement:
\begin{equation}
    \hat{\vec{x}}_0 \;\leftarrow\; \hat{\vec{x}}_0
    - \tau_t \,\nabla_{\hat{\vec{x}}_0}\,\bigl\|A\,\hat{\vec{x}}_0-\vec{y}\bigr\|^2,
\end{equation}
where $\tau_t$ is the (possibly time‐dependent) step size. Finally, we re‐noise the updated $\hat{\vec{x}}_0$ according to the DDIM schedule to produce the next state $\vec{x}_{t-1}$.

In contrast to DPS, which indirectly adjusts $\vec{x}_0$ via small gradients in the noisy $\vec{x}_t$ space, our method explicitly estimates and corrects $\vec{x}_0$ before feeding it back into the diffusion trajectory. This yields a stronger pull toward solutions that more strictly satisfy the measurement constraint $A\vec{x}=\vec{y}$. Pseudocode is given in Algorithm~\ref{algo:DPS}.

\subsection{Differentiable Forward Model}
We make the forward rendering depth–differentiable by instantiating a per-pixel PSF via an affine re-scaling of a calibrated reference coded PSF. Concretely, we calibrate a single reference PSF once, and at each image pixel we generate its PSF by applying an affine transform whose \textit{scale parameter} is a smooth function of the local depth. 
Critically, since gradients must be computed at every pixel for realistic image resolutions, a full wave-optics propagation model would be computationally prohibitive. 
The affine approximation is therefore indispensable: it preserves the essential depth dependence while enabling efficient per-pixel gradient back-propagation, without resorting to depth-plane discretization. 
Implementation uses standard \texttt{im2col/col2im} primitives; explicit formulas are provided in the supplementary material.

\section{Experimental Setup}
This section describes the hardware and simulation settings used in our real-capture and synthetic experiments.

\subsection{Optical Configuration}
\begin{figure}
    \centering
    \includegraphics[width=1\linewidth]{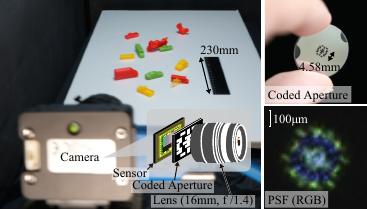}
    \caption{Experimental setup with a prototype camera, a fabricated coded aperture, and a measured PSF. The PSF is normalized per RGB channel and shown in color.}
    \label{fig:experiment_setup}
\end{figure}

Our method does not assume any particular aperture pattern or lens design. As a representative configuration, we follow the optical setup of Levin \textit{et~al.}~\cite{Levin2007}. We employ a 35\,mm focal length with an $F$-number of f/1.8 and a sensor pitch of $13\,\mu$m, adopting the original coded-aperture pattern. For the real-capture experiments, we build a prototype camera with a 16\,mm f/1.4 lens, while retaining the same sensor pitch ($13\,\mu$m) and coded-aperture pattern. The coded aperture is fabricated by depositing chromium on a glass substrate, leaving only a $4.58$\,mm~$\times$~$4.58$\,mm transmissive square window (coded pattern), with all other regions blocked.
Fig.~\ref{fig:experiment_setup} shows the fabricated coded aperture and experimental setup.

The imaging sensor is an RGB CMOS camera with a native resolution of $1536\times 2048$\,px (Teledyne FLIR CM3-U3-13Y3C-CS). The captured images are averaged down by a factor of two and center-cropped to $512\times 512$\,px for reconstruction.

For the simulations, objects are randomly placed at distances of 2--4\,m and the lens focus is set to 1.5\,m. For the real experiments, we restrict the working distance to 0.95-–1.45\,m and fix the focus at 0.62\,m.

PSF calibration uses a white LED source (SIGMAKOKI SLA-100B) and a 200\,$\mu$m pinhole placed at 1.50\,m. This reference PSF is resized and reused both for reconstruction on real data and as the forward model in training. For the simulations, we use the dataset released by Levin \textit{et~al.}

We maintain the PSFs in RGB and, after depth-dependent resizing, renormalize each channel so that its spatial sum equals one. Chromatic aberrations and field dependence are ignored; we assume a shift-invariant, achromatic PSF and apply the \emph{same} depth-to-scale mapping to all three channels by affine rescaling of the reference PSF. Two-dimensional convolutions are applied independently per channel and the rendered observation is composed in RGB.

\subsection{Datasets}
To demonstrate that our approach operates without entangling the camera parameters with the learning process, we align the synthetic data as closely as possible to the real setup. We generate scenes with \textsc{superCLEVR}~\cite{li2023super} via 3D computer graphics software~(Blender), adjusting the focal length and object placement so that the targets lie at distances of 2--4\,m. To avoid a degenerate planar floor at constant depth, we randomize the camera pose (position and tilt), inducing a random floor slope.

Under these conditions, we render 15,000 RGBD images for training and 100 images for testing; the test set is constructed with additional object models not present in the training set. To better emulate real captures, we add i.i.d.\ Gaussian noise during both training and inference, with signal-to-noise ratios of $\sigma = $ 0.0316(30\,dB), 0.01(40\,dB), and 0.00316(50\,dB).

For the real experiments, the synthetic distribution is mirrored by staging scenes on a matte white board and randomly arranging 3D-printed toys.   

\section{Simulations}
We first validate the proposed method through a series of simulations.

\subsection{Baselines}
There has been limited work on estimating depth from a single-snapshot coded-aperture DFD measurement. Among the methods with publicly available code or sufficiently detailed procedural descriptions, we compare the following methods.

\textbf{Levin \textit{et al.}}~\cite{Levin2007}\;
This was the first demonstration of coded‐aperture DFD. Depth and the all‐in‐focus (AIF) image are jointly recovered by minimizing a data‐fidelity term together with first‐ and second‐order derivative norms. We use this classical, non–deep learning approach as a baseline. 
The original work additionally included post-processing steps (e.g., graph cuts with user scribbles) to refine the visual quality of the output. 
However, these steps are not part of the core reconstruction algorithm, and the corresponding implementations are not publicly released. 
Accordingly, we focus our comparisons on the main reconstruction pipeline without such auxiliary refinements.

\textbf{DIP}~\cite{Ulyanov2020}\;
While Deep Image Prior (DIP) strictly refers to a regularization strategy rather than a reconstruction algorithm itself, for simplicity we denote this baseline as DIP. 
In this approach, random noise is fed into an untrained CNN (typically a U-Net), which is optimized so that its output matches the target imag reconstruction. 
The network architecture itself acts as an implicit regularizer, suppressing noise and guiding the reconstruction. 
Although DIP has not been previously applied to DFD, we include it as an intermediate baseline between Levin's classical optimization and our proposed diffusion-prior framework, highlighting the role of learned priors without requiring external training data.

\textbf{Baek \textit{et al.}}~\cite{Baek2022} \textbf{/ Wu \textit{et al.}}~\cite{Wu2019}\;
These methods are recent approaches that learn a phase/coded mask and use a U‐Net–style reconstruction network to regress the depth and AIF image. While they differ in mask optimization and training losses, the reconstruction paradigm (U‐Net regression from defocus) is common.

In our study, the mask is fixed (no co‐design); we update only the depth–reconstruction algorithm. Accordingly, we compare against Levin's optimization‐based method, DIP and the U‐Net reconstruction baselines (Baek/Wu) under a shared coded aperture. 
Hyperparameters for Levin and DIP are selected using a validation set. 
The U‐Net baselines are trained following the original papers, including their training schedules and hyperparameters.

\begin{table*}[t]
    \centering
    \caption{Average over 100 simulated test scenes. Metrics: depth mean absolute error (MAE; $\rm m$) and AIF peak signal-to-noise ratio (PSNR; $\rm dB$).}
    \label{tab:train_eval_snr_grid}
    \begin{tabular}{l c| *{3}{c} | *{3}{c}} 
      \toprule
      & &
      \multicolumn{3}{c}{\makecell{\textbf{Depth (MAE $\downarrow$) [m]}\\ \scriptsize Eval SNR}} &
      \multicolumn{3}{c}{\makecell{\textbf{AIF (PSNR $\uparrow$)}\\ \scriptsize Eval SNR}} \\
      \cmidrule(lr){3-5}\cmidrule(lr){6-8}
      \textbf{Method} & \textbf{Train SNR} & $\sigma$=0.0316 & $\sigma$=0.01 & $\sigma$=0.00316 & $\sigma$=0.0316 & $\sigma$=0.01 & $\sigma$=0.00316 \\
      \midrule
      \multirow{3}{*}{Wu \etal~\cite{Wu2019}}
          & \bf{$\sigma$=0.0316} & \bfhl{0.025} & 0.041 & 0.053      & \bfhl{36.17} & 36.09 & 36.03 \\ 
          & \bf{$\sigma$=0.01} & 0.141 & \bfhl{0.018} & \hl{0.023} & 30.19 & \bfhl{39.26} & 39.52 \\ 
          & \bf{$\sigma$=0.00316} & 0.504 & 0.079 & \bfhl{0.015}      & 28.97 & 36.40 & \bfhl{41.10} \\ 
      \addlinespace
      \multirow{3}{*}{Baek \etal~\cite{Baek2022}}
          & \bf{$\sigma$=0.0316} & \bfhl{0.020} & 0.031 & 0.038      & \bfhl{37.41} & \hl{38.00} & 37.97 \\
          & \bf{$\sigma$=0.01} & 0.095 & \bfhl{0.020} & 0.030      & 27.68 & \bfhl{39.46} & 39.75 \\
          & \bf{$\sigma$=0.00316} & 0.316 & 0.039 & \bfhl{0.013}      & 24.85 & 34.83 & \bfhl{41.25} \\
      \addlinespace
      Ours & --     & 0.028 & 0.025 & 0.025            & 34.38 & 37.38 & 40.75 \\
      \addlinespace
      DPS \cite{chung2023diffusion,chung2023parallel} 
           & --     & 0.679 & 0.681 & 0.682 & 26.33 & 26.64 & 26.67\\
      Levin \cite{Levin2007} 
           & --     & 0.854 & 0.772 & 0.685 & 23.58 & 28.37 & 28.40\\
      DIP \cite{Ulyanov2020}  &  --  & 0.533 & 0.516 & 0.519 & 32.40 & 32.54 & 32.56\\
      \bottomrule
    \end{tabular}
\end{table*}
\begin{figure*}
    \centering
    \includegraphics[width=0.9\linewidth]{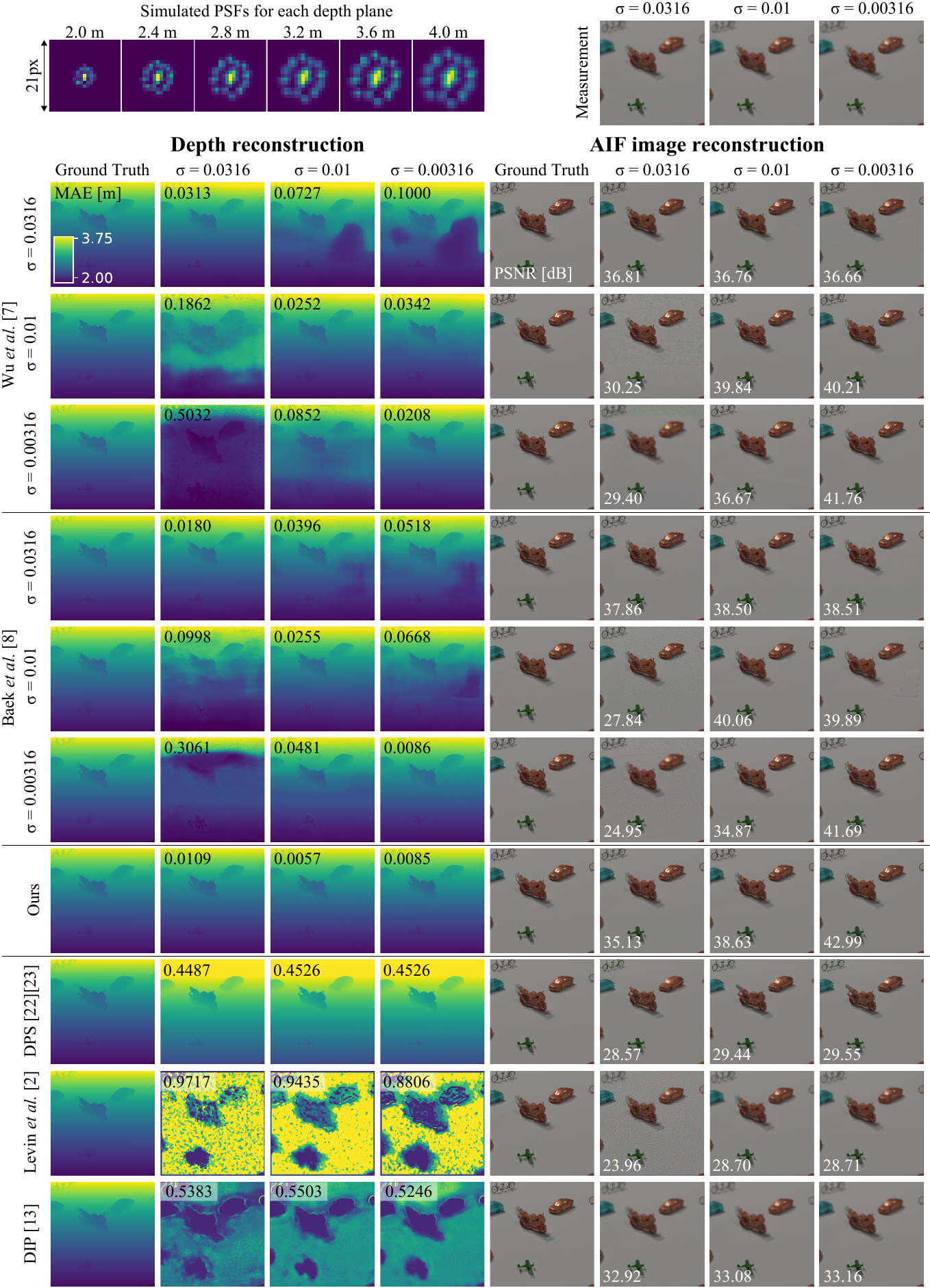}
    \caption{Simulated PSFs, observed images, and reconstruction results. We show the U-Net baselines, the proposed method, DPS, Levin \textit{et~al.}, DIP.  For the U-Net baselines, the results are arranged with the training SNR in rows and the evaluation SNR in columns.}
    \label{fig:sim_75}
\end{figure*}

\begin{figure}
    \centering
    \includegraphics[width=1\linewidth]{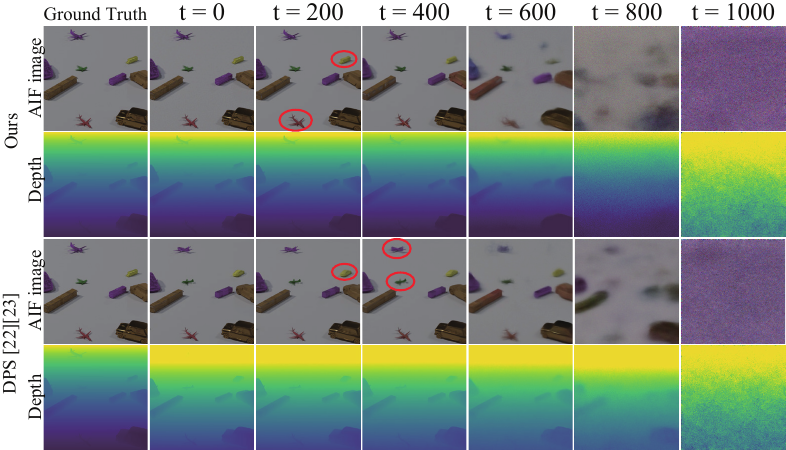}
    \caption{Comparison of generation trajectories for the proposed method and DPS. DPS exhibits hallucinations, whereas the proposed method reconstructs without such hallucinations by enforcing measurement consistency in the denoised $\hat{\vec{x}}_0$ space.}
    \label{fig:generation_process}
\end{figure}

\subsection{Results and Discussion}
We evaluate both the average performance over 100 test scenes (Table~\ref{tab:train_eval_snr_grid}) and representative examples (Fig.~\ref{fig:sim_75}). The evaluation metrics are the depth mean absolute error (MAE) and the AIF image peak signal-to-noise ratio (PSNR). The U-Net baselines (Wu \etal, Baek \etal) are compared on a square grid with training SNRs as rows and evaluation SNRs as columns; for the proposed method, DPS, and Levin---none of which requires training tied to the observation SNR---the row entry is marked ``--''.

\paragraph{Quantitative results (Table~\ref{tab:train_eval_snr_grid}).}
The proposed method attains stable accuracy across all evaluation SNRs and, in particular, significantly outperforms the U-Net baselines when the training and evaluation SNRs are mismatched. The U-Net baselines achieve their best scores on the Train{=}Eval diagonal, but degrade markedly off-diagonal. This pattern holds for both the depth MAE and AIF PSNR, supporting the robustness of our approach in which the physical imaging parameters are decoupled from learning. DPS produces large errors at all SNRs. Levin~\etal underperform because of its texture dependence and the noise amplification inherent to deconvolution, which limits the PSNR. 
DIP shows improved performance compared to Levin~\etal, particularly in textureless regions where the implicit regularization of the untrained network acts effectively. However, this regularization essentially operates as value interpolation, leading to superficially high reconstruction scores. Similar to Levin~\etal, DIP does not leverage semantic or contextual cues from the AIF image to infer depth, and thus the accuracy of the recovered depth maps remains limited.

\paragraph{Qualitative results (Fig.~\ref{fig:sim_75}).}
Our reconstructions preserve shape and texture while satisfying the measurement model. In contrast, the U-Net baselines exhibit artifacts in AIF where depth estimation fails, visually confirming their strong dependence on the training conditions. DIP yields smoother results than Levin~\etal in textureless regions, but the reconstructions lack meaningful depth cues, reflecting its reliance on interpolation-like regularization rather than true depth reasoning.

\subsection{Hallucination Analysis}\label{sec:hallucination}
We further analyze the generation trajectories in Fig.~\ref{fig:generation_process}. DPS performs only small updates in the noisy $\vec{x}_t$ space; although object-like structures appear early, the absolute depth drifts and object category/orientation swaps (hallucinations) persist. 
In contrast, our method explicitly estimates $\hat{\vec{x}}_0$ at each step, enforces measurement consistency via
$\nabla_{\hat{\vec{x}}_0}\!\left\lVert A\hat{\vec{x}}_0-\vec{y}\right\rVert^2$
in the denoised space, and then re-noises. 
This $\vec{x}_0$-space correction progressively reduces the measurement error while suppressing semantic failures.

\section{Experiments}

\begin{figure}
    \centering
    \includegraphics[width=1\linewidth]{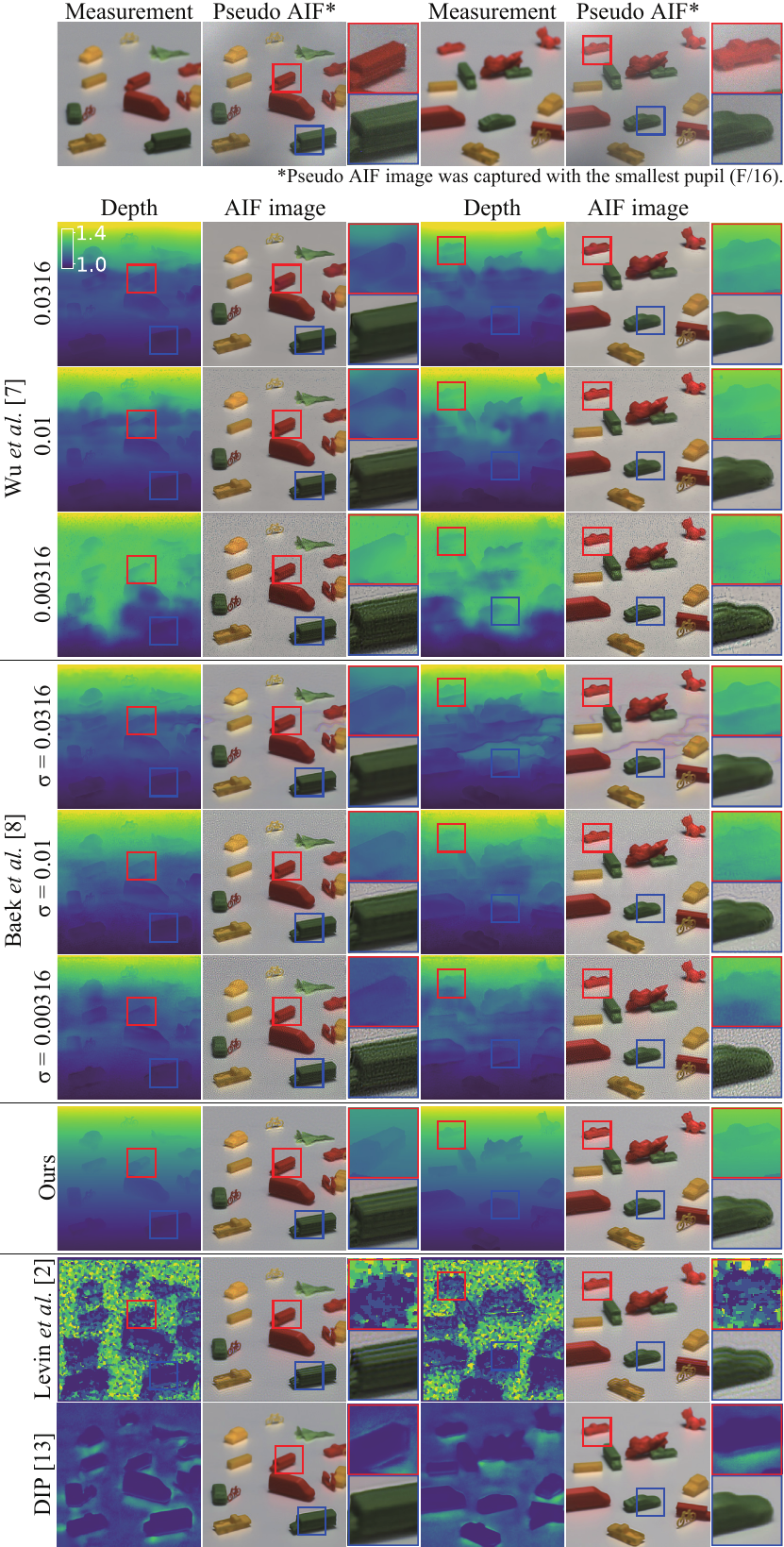}
    \caption{Reconstruction results on real captures. Top: coded-aperture observation and a pseudo all-in-focus (AIF) image captured with the aperture fully stopped down (largest f-number). Bottom: reconstructions of each method. Red and blue boxes indicate zoomed-in regions.}
    \label{fig:real_results}
\end{figure}

In this section, we validate the proposed method on real measurement data acquired with a prototype camera.

\subsection{Baselines}
As in the simulation study, we compare against representative U-Net-based reconstructions (Baek \textit{et al.}~\cite{Baek2022}, Wu \textit{et al.}~\cite{Wu2019}) and the optimization-based method of Levin \textit{et al.}~\cite{Levin2007} and DIP. Because the U-Net baselines require training, instead of collecting defocus--RGBD pairs in the wild, we synthesize defocus training images by combining the PSF calibrated on our physical camera with the RGBD dataset used in the simulations, after correcting its depth distribution to match the real setup. The training noise conditions are identical to those of the simulations ($\sigma = 0.0316, 0.01, 0.00316$), and each method's hyperparameters and training schedules follow their respective papers.

\subsection{Results}
Fig.~\ref{fig:real_results} shows, for two scenes, the coded-aperture observation, the pseudo-AIF image captured at the minimum f-number~($\rm{F}/16$), and the reconstructions of all methods. As the ground truth is unavailable for real data, we use the pseudo-AIF as a visual reference.

Across models, the U-Net baselines exhibit depth artifacts similar to those observed in the simulations when the training and evaluation conditions diverge. In the AIF results, models trained with larger added noise show attenuated high-frequency edges, illustrating their dependence on the training conditions.

Overall, our method preserves shape and texture while remaining consistent with the measurement model. In the left red box, a shallow concavity appears inside the orange bus; this likely reflects the influence of regularization in a texture-poor region. By contrast, in the right red box, the recess at the rear of the truck is correctly reproduced. For the AIF images, as in the simulations, some slight noise amplification is attributable to deconvolution, but there are no evident hallucinations. In the left blue box, the windows of the green bus are preserved by our method, whereas they collapse for the U-Net baselines. This area is also reconstructed by the method of Levin \textit{et al.}, suggesting that the gradient from the data term is functioning effectively.

Levin's method yields relatively good depth near object boundaries, but reconstruction is degraded in texture-poor regions because of the weak regularization. Deconvolution-induced noise amplification is pronounced, which limits the AIF image quality.
DIP, similar to Levin's method, fails to recover accurate depth away from edges. 
This is consistent with the AIF reconstructions, which lack high-frequency details, indicating that defocus components are not properly disentangled. 
These observations suggest that the implicit regularization of DIP is insufficient for robust depth recovery.
\section{Conclusion and Future Work}
We have presented a single–image DFD method that combines a coded aperture with a diffusion-model prior.
Our pipeline (i) builds a differentiable forward model for coded apertures by instantiating per-pixel PSFs via depth-to-scale affine re-scaling and (ii) specializes DPS to DFD by performing a denoise--likelihood--re-noise update in the $\vec{x}_0$ space. Crucially, the diffusion model is used only as a regularizer, decoupling camera calibration from network training and enabling reconstruction without paired defocus--RGBD data.

Across synthetic benchmarks, the proposed method delivered stable, high‐quality reconstructions regardless of the evaluation settings. On real captures with our prototype camera, it likewise produced RGBD reconstructions that adhered to the observation model more faithfully than the baselines. These findings support our design choice: enforcing data fidelity in the denoised $\vec{x}_0$ space yields stronger measurement consistency than applying likelihood gradients in the noisy $\vec{x}_t$ domain.

\paragraph{Future work.}
We see three immediate directions. (i) Coded-aperture design for DFD--DPS: although our method is not end-to-end trainable with mask parameters as in U-Net regressors, we plan to co-design aperture patterns offline by optimizing a differentiable proxy for the reconstruction error under our $x_0$-space update scheme. (ii) Failure-mode analysis on real data: we will systematically study when and why artifacts such as spurious concavities in texture-poor regions arise, disentangling the roles of noise level, PSF miscalibration, and forward-model mismatch, and exploring remedies (e.g., confidence-aware updates and stronger depth priors). (iii) Scaling the prior: we will pretrain on substantially larger RGBD datasets to build a more scene-independent diffusion prior and assess transfer to out-of-distribution scenes, aiming for a ``foundation'' RGBD model that is usable across diverse settings.

\bibliographystyle{IEEEtran}
\bibliography{main}

\vspace{11pt}

 \begin{IEEEbiography}[{\includegraphics[width=1in,height=1.25in,clip,keepaspectratio]{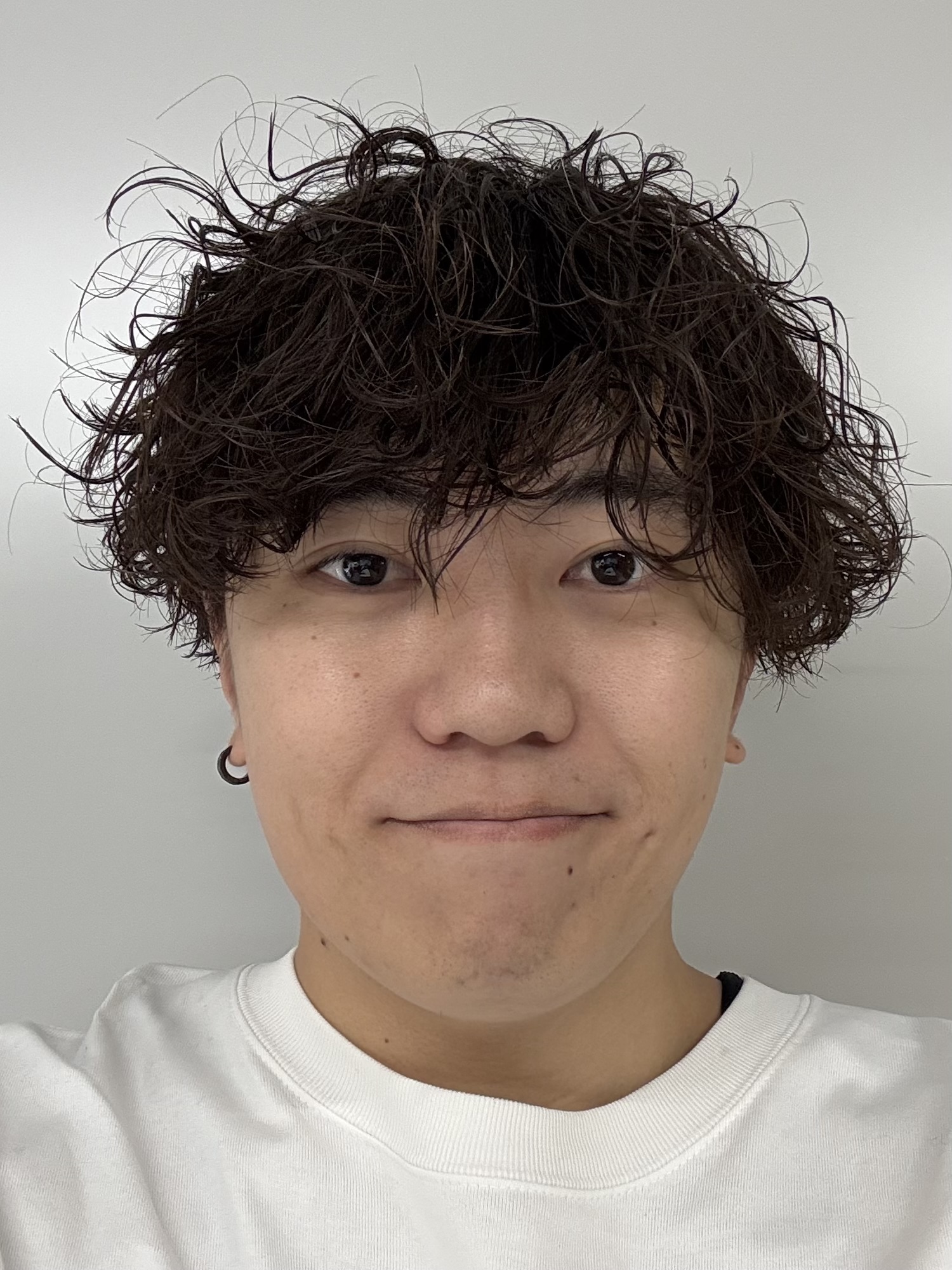}}]{Hodaka Kawachi}
 received the B.S. in engineering and M.S. in computer science from Osaka University, in 2022 and 2024 respectively. Currently, he is a PhD candidate on the computer science department of Osaka University. His research interests include computational photography, with a preference for depth imaging, and deep learning techniques applied to computer vision.
 He has been awarded several honors, including the 27th Meeting on Image Recognition and Understanding (MIRU2024) MIRU Student Award (2024), the Graduate School of Information Science and Technology Award from Osaka University (2023), the 13th Japan-Korea Workshop on Digital Holography and Information Photonics (DHIP2023) Student Presentation Award (2023), the 25th Meeting on Image Recognition and Understanding (MIRU2022) MIRU Student Award (2022), and the Konica Minolta Optics Future Encouragement Award (2024). He has also been selected as a recipient of the JSPS DC1 fellowship since 2024.
 \end{IEEEbiography}

\begin{IEEEbiography}[{\includegraphics[width=1in,height=1.25in,clip,keepaspectratio]{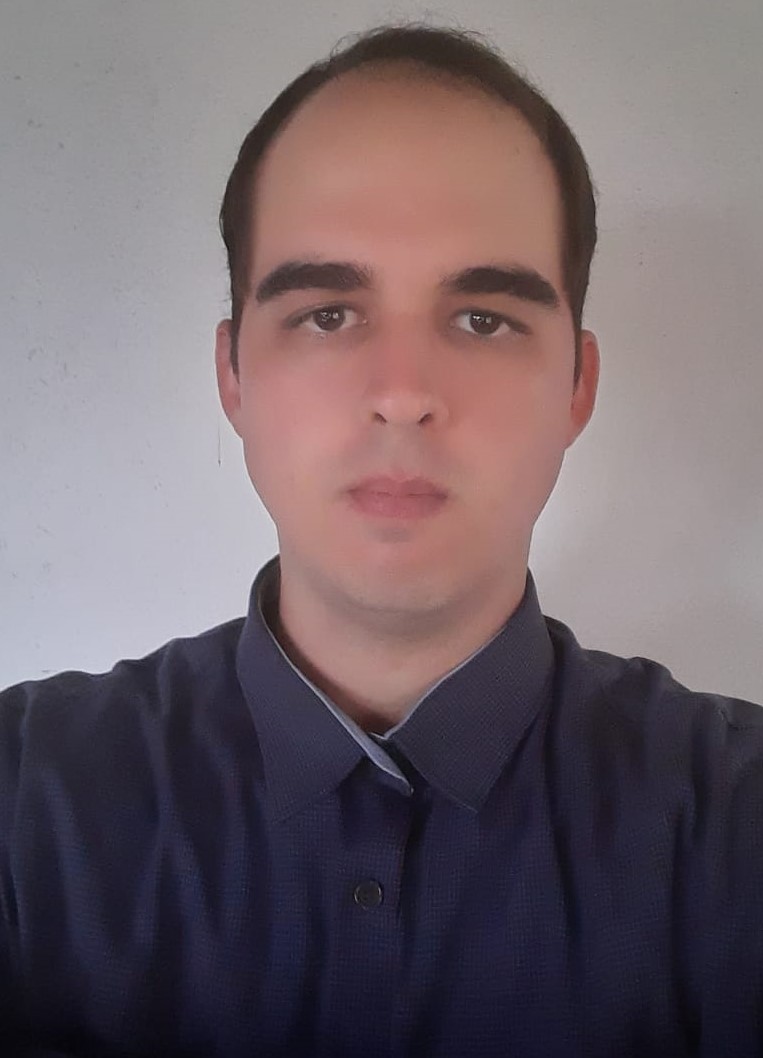}}]{Jos\'{e} Reinaldo Cunha Santos A. V. Silva Neto}
received the B.S. in engineering and M.S. in computer science from the University of Brasilia, in 2019 and 2021 respectively. Currently, he is a PhD candidate on the computer science department of Osaka University. His research interests include computational photography, with a preference for lensless imaging, and deep learning techniques applied to computer vision. He is a member of the optical society of Japan, and received awards such as the 13th Japan-Korea Workshop on Digital Holography and Information Photonics (DHIP2023) Student Presentation Award (2023).
\end{IEEEbiography}
 
 \begin{IEEEbiography}[{\includegraphics[width=1in,height=1.25in,clip,keepaspectratio]{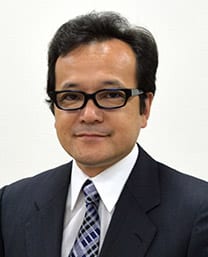}}]{Yasushi Yagi}
(Senior Member, IEEE) received the Ph.D. degree from Osaka University, in 1991. In 1985, he joined the Product Development Laboratory, Mitsubishi Electric Corporation, where he was involved in robotics and inspections. He became a Research Associate, in 1990, a Lecturer, in 1993, an Associate Professor, in 1996, and a Professor, in 2003, with Osaka University, where he was the Director of SANKEN (The Institute of Scientific and Industrial Research), from 2012 to 2015. He was the Executive Vice President of Osaka University, from 2015 to 2019. His research interests include computer vision, pattern recognition, biometrics, human sensing, medical engineering, and robotics. He is a fellow of the IPSJ and a member of the IEICE and the RSJ. He is a member of the Editorial Board of the {\it International Journal of Computer Vision}. He is the Vice President of the Asian Federation of Computer Vision Societies. He was awarded the ACM VRST2003 Honorable Mention Award, the IEEE ROBIO2006 Finalist of the T. J. Tan Best Paper in Robotics, the IEEE ICRA2008 Finalist for the Best Vision Paper, the PSIVT2010 Best Paper Award, the MIRU2008 Nagao Award, the IEEE ICCP2013 Honorable Mention Award, the MVA2013 Best Poster Award, the IWBF2014 IAPR Best Paper Award, and the {\it IPSJ Transactions on Computer Vision and Applications} Outstanding Paper Award (2011 and 2013). International conferences for which he has served as the Chair include ROBIO2006 (PC), ACCV (2007PC and 2009GC), PSVIT2009 (FC), and ACPR (2011PC, 2013GC, 2021GC, and 2023GC). He has al
so served as an Editor for the IEEE ICRA Conference Editorial Board (2008 and 2011). He was the Editor-in-Chief of the {\it IPSJ Transactions on Computer Vision and Applications}.
\end{IEEEbiography}

\begin{IEEEbiography}[{\includegraphics[width=1in,height=1.25in,clip,keepaspectratio]{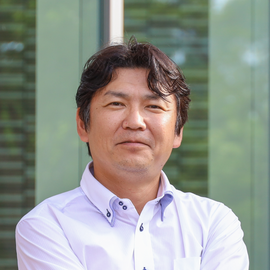}}]{Hajime Nagahara}
Hajime Nagahara is a professor at Institute for Datability Science, Osaka University, since 2017.  He received Ph.D. degree in system engineering from Osaka University in 2001. He was a research associate of the Japan Society for the Promotion of Science from 2001 to 2003. He was an assistant professor at the Graduate School of Engineering Science, Osaka University, Japan from 2003 to 2010.  He was an associate professor in Faculty of information science and electrical engineering at Kyushu University from 2010 to 2017.
He was a visiting associate professor at CREA University of Picardie Jules Verns, France, in 2005. He was a visiting researcher at Columbia University in 2007-2008 and 2016-2017.
Computational photography and computer vision are his research areas. He received an ACM VRST2003 Honorable Mention Award in 2003, IPSJ Nagao Special Researcher Award in 2012, ICCP2016 Best Paper Runners-up, and SSII Takagi Award in 2016. He is a program chair in ICCP2019, Associate Editor for IEEE Transaction on Computational Imaging in 2019-2022, and Director of Information Processing Society of Japan in 2022-2024.
\end{IEEEbiography}
 
 \begin{IEEEbiography}[{\includegraphics[width=1in,height=1.25in,clip,keepaspectratio]{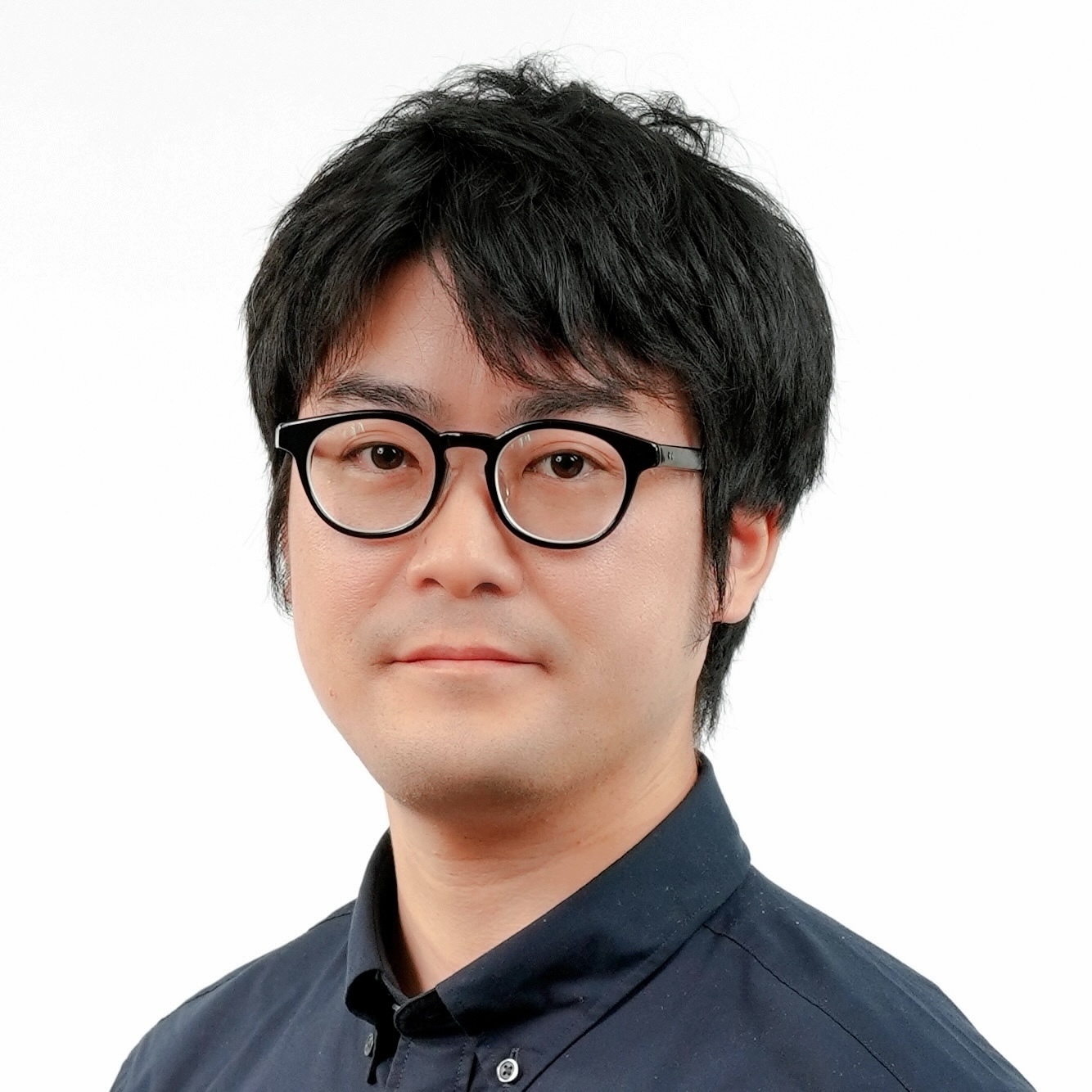}}]{Tomoya Nakamura}
Tomoya Nakamura received the Ph.D. degree from Osaka University, Suita, Japan, in 2015. He is currently an Associate Professor with the Graduate School of Engineering Science, The University of Osaka. His research interests include computational imaging and holography. He is a member of the Optica. He was the recipient of several honors and awards, including the International Display Workshops (IDW2017), the Best Paper Award, the 4th International Workshop on Image Sensor and Systems (IWISS 2018), the Open Poster Session Award first Place, and The 1st Kyowa Technologies ICT Encouragement Award.
 \end{IEEEbiography}
\vfill

\end{document}